\documentclass{article}
\usepackage{spconf,amsmath,graphicx}
\usepackage{caption}
\usepackage{booktabs,multirow}
\usepackage{xcolor}
\usepackage{tikz,pgfplots}
\usepackage[numbers,sort&compress]{natbib}

\newcommand\heading[1]{\par\noindent\textbf{#1}:}

\title{RNN BASED INCREMENTAL ONLINE SPOKEN LANGUAGE UNDERSTANDING}
\name{Prashanth Gurunath Shivakumar$^1$\thanks{\texttt{$^1$email:pgurunat@usc.edu}}, Naveen Kumar$^2$, Panayiotis Georgiou$^1$, Shrikanth Narayanan$^1$}
\address{$^1$University of Southern California, Los Angeles, CA, USA\\
$^2$Disney Research, Glendale, CA, USA}
\begin{document}
\ninept
\maketitle
\begin{abstract}
Spoken Language Understanding (SLU) typically comprises of an automatic speech recognition (ASR) followed by a natural language understanding (NLU) module.
The two modules process signals in a blocking sequential fashion, i.e., the NLU often has to wait for the ASR to finish processing on an utterance basis, potentially leading to high latencies that render the spoken interaction less natural.
In this paper, we propose recurrent neural network (RNN) based incremental processing towards the SLU task of intent detection.
The proposed methodology offers lower latencies than a typical SLU system, without any significant reduction in system accuracy.
We introduce and analyze different recurrent neural network architectures for incremental and online processing of the ASR transcripts and compare it to the existing offline systems.
A lexical End-of-Sentence (EOS) detector is proposed for segmenting the stream of transcript into sentences for intent classification.
Intent detection experiments are conducted on benchmark ATIS, Snips and Facebook's multilingual task oriented dialog datasets modified to emulate a continuous incremental stream of words with no utterance demarcation.
We also analyze the prospects of early intent detection, before EOS, with our proposed system.
\end{abstract}
\begin{keywords}
Incremental Processing, Online Processing, Spoken Language Understanding, Intent Detection, Recurrent Neural Network
\end{keywords}
\section{Introduction}
\label{sec:intro}

With the proliferation of novel interactive technology applications across domains ranging from entertainment to health, the use of spoken language for enabling and supporting natural communication is becoming ever more important.
Today, SLU finds applications in voice assistants, robot interactions, virtual agents, virtual \& augmented reality applications as well as in mediating human interactions such as meetings.
While the rapid development in SLU techniques has led to a revolution in this field, a lot still remains to be bridged in terms of improving the ``naturalness" of these interactions.
For example, achieving low latencies remains crucial to achieve a sense of ``naturalness'' during conversations.
Higher latencies often result in turn-based disruptive conversations which creates the impression of a transactional interaction \cite{levinson2016turn, stivers2009universals}.

Typical gaps between human-human dyadic turns are of the order of 200 ms \cite{levinson2016turn, stivers2009universals,levinson2015timing, heldner2010pauses}. In contrast, most ASRs rely on Inter Pausal Units (IPU) that are upwards of 500 ms to reliably detect the ``end of utterance". During interactions, this latency is often perceived as computational delay due to speech recognition, whereas in reality this delay can be avoided by use of incremental processing architectures.

Previous works \cite{fink1998incremental,selfridge2011stability,kiss2018incremental} have investigated the stability of results when using incremental hypotheses (partials) from a speech recognition system. 
While these partials can be generated with low latency, their semantic stability for downstream NLU tasks remains challenging. 
Earlier attempts to alleviate this issue have used confidence score \cite{devault2009can, devault2011incremental, traum2012incremental} to assess stability of predictions. 
Some works have relied instead on auxiliary methods to predict turn-taking behavior in an agent \cite{khouzaimi2015turn, skantze2017towards, roddy2018investigating, masumura2017online}.

Most of the research efforts in SLU in the NLP community assume offline NLU processing, i.e., (i) ideal, perfect utterance boundaries, and (ii) error-less transcriptions void of any speech recognition errors \cite{guo2014joint,zhang2016joint,hakkani2016multi,kim2017onenet,liu-lane-2016-joint,Liu+2016,goo2018slot,li2018self,e-etal-2019-novel}.
Leading NLU systems are based on RNN \cite{guo2014joint,zhang2016joint,hakkani2016multi,kim2017onenet}, convolutional neural network (CNN) \cite{li2018self} and sequence-to-sequence architectures \cite{Liu+2016} with attention modeling \cite{Liu+2016,goo2018slot,li2018self}.
Joint modeling of SLU tasks like intent detection, language modeling (LM), slot-filling, and named entity detection are found to be beneficial \cite{guo2014joint,zhang2016joint,hakkani2016multi,kim2017onenet,liu-lane-2016-joint,Liu+2016,goo2018slot,e-etal-2019-novel}.
Character level features have also been proposed to achieve state-of-the-art performance in benchmark tasks \cite{kim2017onenet,li2018self}.
Whereas, research efforts in the speech community assume offline ASR and offline NLU processing, i.e., (i) ideal, perfect utterance boundaries, and (ii) ASR errors in transcriptions \cite{masumura2018neural,Shivakumar2019,schumann2018incorporating,zhu2018robust}.
To handle ASR errors, joint SLU-ASR adaptation \cite{gotab2010online} and joint learning of SLU and ASR error correction \cite{schumann2018incorporating,zhu2018robust} have been studied.
Better feature representations involving acoustic information are beneficial in handling ASR errors for SLU \cite{masumura2018neural,Shivakumar2019}.
Although, there have been a few attempts at end-to-end SLU directly from speech signals, the performance are not quite up-to the standards achieved by the traditional approaches involving ASR and NLU \cite{serdyuk2018towards}.

Few research efforts have tried to incorporate incremental processing on ASR \cite{fink1998incremental,selfridge2011stability,kiss2018incremental}.
Even fewer efforts have been made in the context of ASR incremental processing for SLU.
The authors in \cite{liu-lane-2016-joint} proposed a joint online SLU and LM system using RNN.
Although, their proposed model is capable of outputting the intent class posteriors for each time-step, the posteriors were not used for intent prediction itself, but were fed back to the hidden state of the RNN.
Only the intent at the last time-step of the input sequence is used.~
Moreover, the evaluations were made in an offline fashion assuming each input sequence equals a single sentence and assuming the sentence boundaries are known a-priori.

To the best of our knowledge, there has been no prior work dealing with incremental online SLU employing RNN with evaluations conducted in a truly online sense.
In this work, we setup the online incremental SLU processing along with (i) detection of utterance boundaries, and (ii) presence of erroneous ASR transcriptions.
The proposed system is capable of recognizing intents on an arbitrarily long sequence of words with no sentence or utterance demarcations.


\begin{figure*}[t]
\centering
\includegraphics[width=0.9\textwidth,height=0.33\textwidth]{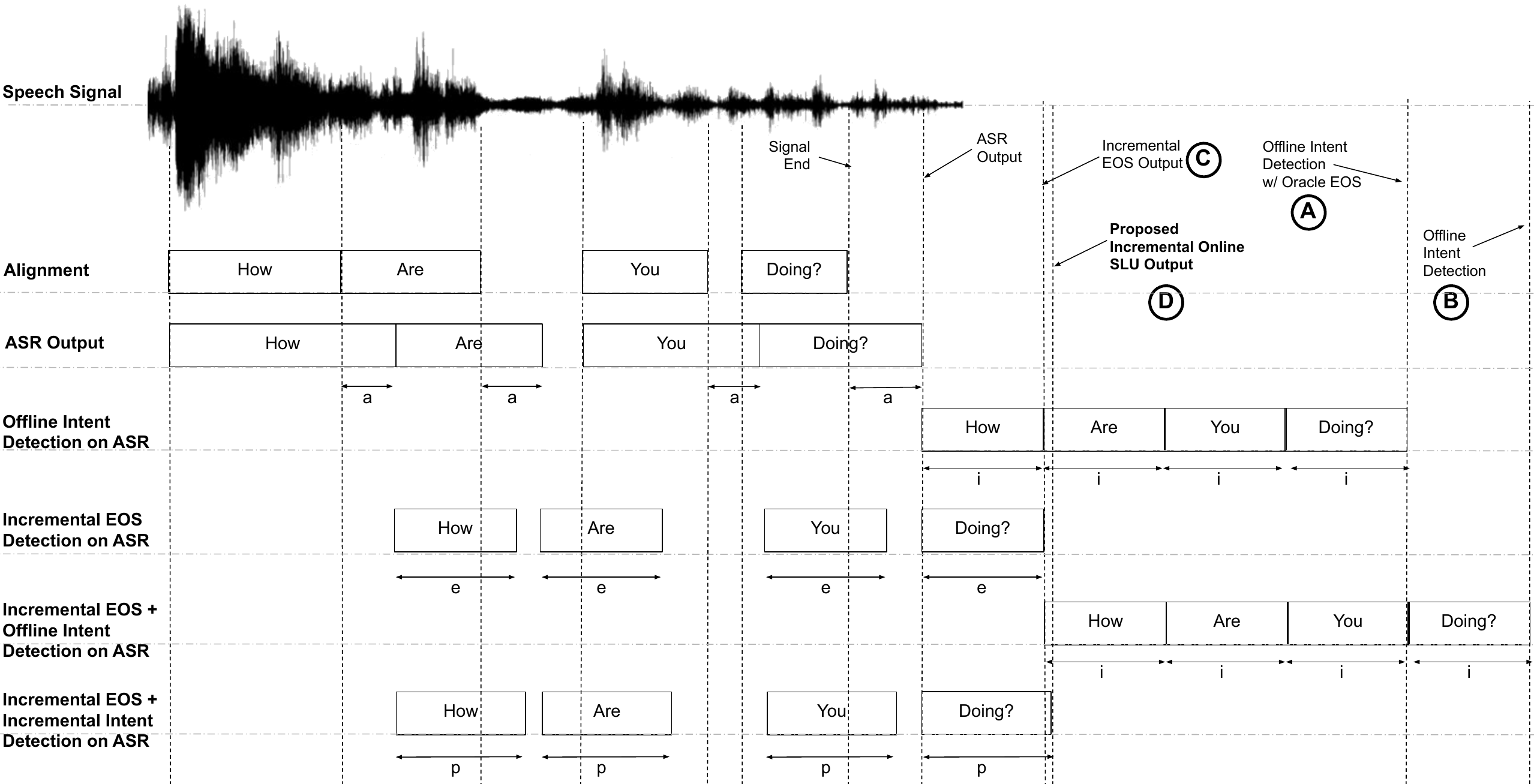}
\caption{Time-line of the entire ASR-SLU pipeline and latency implication of offline versus incremental online systems
(\footnotesize{$a, i, e, p$ are the latencies of ASR, intent classifier, EOS detector, and the latency of the proposed system respectively. $a,i,e,p$ assumed to be constant for simplicity)}}\label{fig:timeline}
\end{figure*}

\section{Need for Incremental Online Processing}
\label{sec:incr_proc}
To motivate the need for incremental online processing, in a real-life, real-time processing system, we present 2 scenarios:

\heading{Scenario 1: Endpoint-based Processing}
In a real-time application scenario, the ASR receives a stream of continuous speech signal and outputs the corresponding transcriptions in real-time.
Due to the computational complexity and memory constraints, most ASRs typically operate by chunking and processing the speech in segments.
This process is often referred to as end-pointing, and is usually determined based on different heuristics related to duration of IPUs, with the goal to minimize disruption during speech.
Finally, the ASR outputs the transcript corresponding to each speech segment.
In this scenario, the ASR is tuned for real-time application, by varying the parameters for end-pointing, often in a heuristic way.
As a result, any application operating on the output of the ASR needs to wait at least until end-pointing, which gives rise to a fundamental bottleneck in latency.

\heading{Scenario 2: Incremental Processing}
Alternatively, during ASR decoding, intermediate querying of ASR output transcript is possible.
This involves computation of the best path over intermediate, incomplete decoded lattices (see example in Fig~\ref{fig:partials}).
Although, there is a possibility of the best path deviating between the complete and incomplete lattice decoding, the deviation is expected to be minimum in robust ASR systems.
Moreover, the prospects of using incremental outputs of the ASR is attractive.
In this scenario, the downstream application has no constraints of waiting until end-pointing and is free to process the ASR transcripts in an incremental manner.
This also allows for online processing for downstream application in addition to the ASR itself for optimal latency.


\subsection{Incremental Processing for SLU Tasks}\label{sec:need_slu}
In the context of SLU, under \emph{Scenario 1}, the NLU module is run in an offline fashion, processing an utterance at each end-point of the ASR.
The timeline is illustrated in Fig~\ref{fig:timeline}.A \&~\ref{fig:timeline}.B.
It is evident from the timeline that offline NLU processing has higher latency implications.
Moreover, the end-pointing algorithm itself has a bearing on the performance of the NLU, since end-pointing defines the utterance boundaries fed to the NLU. 
There have been several research efforts in predicting optimal end-point for an ASR \cite{maas2018combining,liu2015accurate,maas2017domain}.
Thus, the NLU has to deal with the errors from: (i) ASR, (ii) end-point detection, and (iii) ASR errors due to sub-optimal end-pointing (Note, suboptimal end-pointing especially false alarms can result in errors during recognition itself \cite{maas2018combining}). 
The aggregated errors often lead to degradation in the overall performance of SLU.

In this paper, we propose an SLU system under \emph{Scenario 2}, where the NLU module can also be run in an online fashion, in parallel with the ASR.
Additionally, we also propose incremental processing independent of ASR end-pointing and a lexical EOS detection module for utterance boundaries, operating on the ASR output.
This allows for lenient end-pointing schemes (emphasis on lower false positives) since end-pointing no longer defines latency.
This comes with the advantage that the NLU has to deal with errors from only the ASR phase.
However, note that the EOS module might still introduce errors into the system, due to improper segmentation.
The time-line of the incremental processing system is illustrated in the Fig~\ref{fig:timeline}.D.
It is apparent that there are significant latency advantages associated with the proposed system.

\subsection{Implications on Neural Network Architectures}\label{sec:implications}
The online incremental nature of the NLU module imposes certain design constraints on the architecture of the recurrent neural networks.
One of the fundamental restrictions due to the online nature of the problem is the use of only unidirectional LSTM.
This is because, we don't have access to the future time steps for the backward step in-case of bidirectional LSTM.

\section{Proposed Techniques}
\label{sec:proposed}
\subsection{Baseline Offline RNN}\label{sec:baseline}
The baseline system consists of a vanilla RNN-LSTM architecture which consumes a sequence of words and outputs a single decision similar to most of the works \cite{guo2014joint,zhang2016joint,hakkani2016multi,kim2017onenet,liu-lane-2016-joint,Liu+2016,goo2018slot,li2018self} with the exception that the LSTM is unidirectional as per Section~\ref{sec:implications}.
The network is referred to as offline, since the input needs to be segmented such that each utterance has a single intent label during training.
However to assess its performance for the online task, during evaluation, we derive the output per each time step by sharing the output layer over all the time-steps.
The latency implication of baseline is illustrated in Fig~\ref{fig:timeline}.A.

\begin{figure}[t]
\centering
\includegraphics[width=\columnwidth]{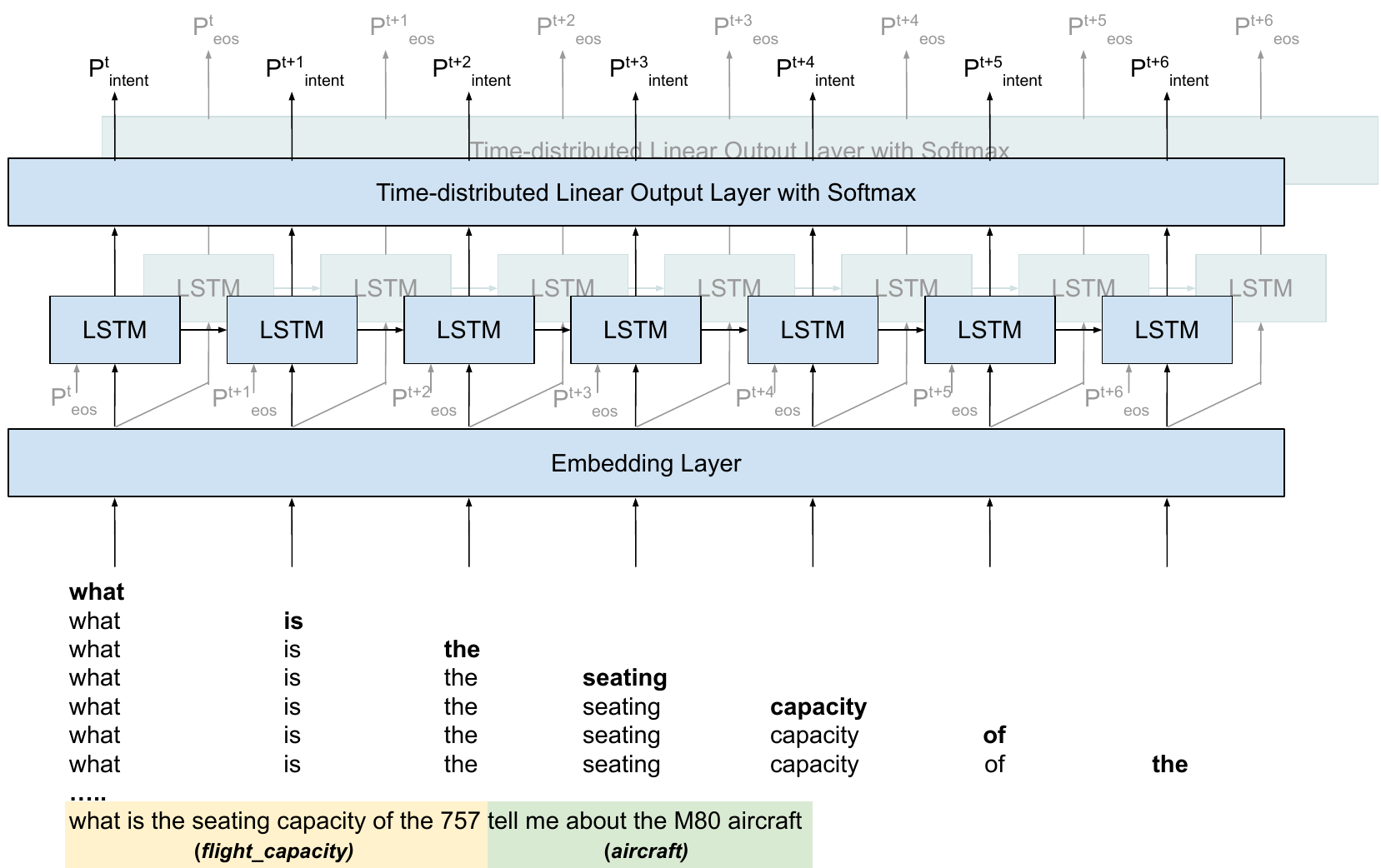}
\caption{Proposed LSTM Architecture (simulation of ASR partials, bold words are increments. 2 utterances, 2 intents)}\label{fig:partials}
\end{figure}

\subsection{Online RNN Classification}\label{sec:baseline_online}
The online version of the system is similar architecture wise to the offline model with the exception that each input time-step has a corresponding output.
The network is referred to as online, since it can process arbitrary length sequences with sequences of multiple intent labels both during training and testing.
The system is trained with input sequences comprising multiple sentences/utterances which possibly map to a sequence of multiple different intents.
During training, since each utterance has a single label, we mask the loss function to compute only at utterance boundaries.
The loss function is:
\begin{equation}\label{eq:mask_loss}
\resizebox{.7\hsize}{!}{$
Loss = - \sum_{t=1}^{T} I_{EOS} \sum_{c=1}^{C} y_{o,c}~log~p(\hat{y}_{o,c})
$}
\end{equation}
where $t$ is the time-step, $T$ is the sequence length, $I_{EOS}$ is the indicator function which is 1 for oracle EOS, $c$ is the intent class, $C$ is the total number of intent classes, $y_{o,c}$ is the indicator function which is 1 if the observation $o$ belongs to class $c$, and $p(\hat{y}_{o,c})$ is the softmax probability prediction.
The latency of the online system is pictured in Fig~\ref{fig:timeline}.D.

\subsection{End-of-Sentence Detection}\label{sec:eos}
Both the baseline Offline and Online RNN system do not have a sense of utterance boundaries during the evaluation phase.
Thus, as described in Section~\ref{sec:need_slu}, we train an independent EOS classifier.
The architecture of the EOS classifier is similar to the online RNN intent classifier with 2 exceptions: (i) sigmoid activation at output, and (ii) binary cross-entropy loss.
In conjunction with the EOS system the latency of baseline offline system and online system is pictured in Fig~\ref{fig:timeline}.B.

\subsection{Online Multi-task Learning}\label{sec:multi_task}
Additionally, we propose to model both the tasks i.e., intent and EOS detection jointly in a multi-task learning framework.
Here, both the tasks share the embedding layer, but have task specific LSTM and time-distributed linear output layer (Fig~\ref{fig:partials} without feedback).
The loss optimized is:%

\begin{equation}\label{eq:multi_loss}
\resizebox{.33\textwidth}{!}{$
\begin{aligned}[]%
Loss = & - \sum_{t=1}^{T} I_{EOS} (\sum_{c=1}^{C} y_{o,c}~log~p(\hat{y}_{o,c})) \\
& + y_{e}~log~p_{e} + (1-y_{e})~log(1-p_{e})
\end{aligned}
$}
\end{equation}
where $y_{e}$ is the oracle EOS label, $p_{e}$ is the predicted output of the network, the rest of the parameters comply with Eq.\ref{eq:mask_loss}.
The proposed system has two advantages: (i) the latency is reduced since both tasks are modeled together, and (ii) joint learning of two tasks can benefit each other as in \cite{guo2014joint,zhang2016joint,hakkani2016multi,kim2017onenet,liu-lane-2016-joint,Liu+2016,goo2018slot}.

\subsection{Online Multi-task with EOS Feedback}\label{sec:multi_task_fb}
Further, within the multi-task learning framework, we experiment with feeding back the EOS output back to the intent detection LSTM.
The embedding layer is shared between the two tasks, with task specific LSTM layers and time-distributed linear output layers.
The predicted EOS output is concatenated along with the input features from the embedding layer and fed to the intent LSTM (see Fig~\ref{fig:partials}).
The loss function is identical to the multi-task learning in equation~\ref{eq:multi_loss}.
With this proposed framework, we believe that explicitly feeding the EOS markers to the intent detection system could provide performance benefits.
The latency is identical to the multi-task learning framework described in section~\ref{sec:multi_task}.


\section{Data \& Experimental Setup}
\label{sec:data_exp_setup}
\subsection{Data}
We employ the ATIS (Airline Travel Information Systems) benchmark dataset \cite{hemphill1990atis} as the primary dataset for the intent detection experiments.
The dataset consists speech recordings of humans speaking to an automated airline travel inquiry system.
The speech recordings are accompanied by manual transcriptions of the spoken queries with annotated intent labels.
The data consists of 17 unique intent categories (in-case of multiple intents, more frequent intent is assigned).
Our data setup is identical to \cite{Shivakumar2019,hakkani2016multi,goo2018slot}.

We also perform experiments on Snips dataset released as a part of the Snips voice platform \cite{snipsdataset} and Facebook's multilingual task oriented dialog dataset \cite{schuster-etal-2019-cross-lingual}.
With Snips we hope to prove the efficacy of the proposed system over multiple domains and extensive vocabulary.
Snips contains 7 intent categories.
The Snips database split is consistent with previous works \cite{goo2018slot,e-etal-2019-novel}.
While ATIS is restricted to a single domain of air travel, Snips involves data spanning multiple domains.
Snips is also characterized by a much larger vocabulary (11420 words) compared to ATIS (869 words).

With Facebook's multilingual task oriented dialog (FMTOD), we explore the transferability of the proposed system over multiple languages.
FMTOD comprises of data spanning three languages, English, Spanish and Thai database.
Each language contains 12 intent categories spanning 3 domains.
The database split is consistent with \cite{schuster-etal-2019-cross-lingual}.

\subsection{Experimental Setup}
To simulate online continuous stream of transcriptions, random number of samples from manual transcripts of the ATIS dataset were stitched together without exceeding a maximum utterances limit (example in Fig~\ref{fig:partials}).
This results in each sample containing multiple utterances with sequence of multiple intent labels with no demarcation.
Instead of random sampling, we also tried concatenating utterances in sequence to incorporate any dialog flow.
However, we found no significant difference in performance. 
We believe the nature of the database with independent utterance structures provide no significant dialog flow. 
Hence random sampling is adopted.
We create multiple copies of each of the dataset by varying maximum number of utterances from 1-10 for analysis.
The data contains exactly the same information and is consistent with previous studies involving ATIS \cite{Shivakumar2019,hakkani2016multi,goo2018slot} facilitating direct comparisons. 

The RNN-LSTM models are trained on the samples from the training set and the development set is used for hyper-parameter tuning.
Finally, the model with the best performance on the development set is chosen and evaluated on the unseen, held out test set.
A single layer LSTM model is adopted with the embedding layer dimension set to 556 taking recommendations from ~\cite{Shivakumar2019}.
The hidden dimension of LSTM was tuned over {32, 64, 128, 256} units.
The dropout is varied over {0.1, 0.15, 0.20, 0.25, 0.3}.
The batch size is set to 1 with each sequence of utterances viewed as a single sample.
The learning rate of 0.001 is used along with the Adam optimizer and trained for a total of 20 epochs.
The convergence of the model is ensured by examining the loss and classification accuracies.

For performing experiments on the speech recognition output, we choose the ASpIRE model\footnote{http://kaldi-asr.org/models/m1} offered by Kaldi open source ASR toolkit \cite{povey2011kaldi}. 
The ASpIRE model is trained Fisher English Corpus \cite{cieri2004fisher} by augmenting the speech corpora using reverberation and noise data.
The acoustic model is based on the time-delay neural network (TDNN) architecture which employs sub-sampling of the outputs to reduce complexity \cite{peddinti2015jhu}.
The language model is a simple tri-gram model trained on the Fisher English training corpus.
An offline decoding of ASR is performed on the test split of the ATIS dataset which yielded a word error rate (WER) of 15.89\%.

\begin{figure}[t]
\centering
\includegraphics[width=\columnwidth]{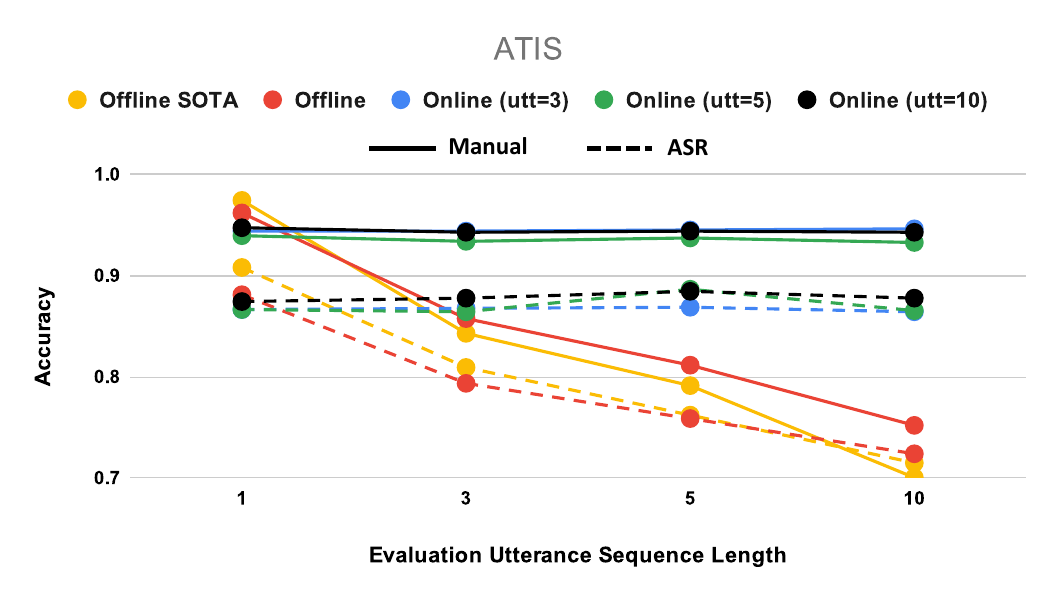}
\caption{\small{Accuracy of offline, SOTA vs. proposed online models at oracle EOS on Manual and ASR ATIS transcripts.}}\label{fig:pre_results}
\end{figure}

\section{Results}
\label{sec:results}


\subsection{Offline Model vs. Proposed Online Model}\label{sec:results_atis}
We first validate the effectiveness of the proposed online model described in section~\ref{sec:baseline_online} against the baseline offline model (Section~\ref{sec:baseline}) as well as the state-of-the-art (SOTA) in offline intent detection \cite{e-etal-2019-novel}.
Fig~\ref{fig:pre_results} plots the results, i.e., Accuracy of the baseline offline model versus the proposed online models over varying utterance lengths.
Three versions of the proposed online models are trained and evaluated with combinations of varying length of utterance sequences ($1, 3, 5, 10$).
The oracle end-of-utterance is assumed during the evaluation.
From the plot, we observe that the accuracy of the baseline offline model as well as the SOTA model is maximum for offline decoding (utterance sequence length of 1) and drops with the increase in utterance sequence lengths.
Whereas, the performance of the online models is slightly lower for utterance sequence of length of 1 compared to the baseline and SOTA model, but doesn't degrade with increasing length of utterance sequences.
In the case of the SOTA model, we observe that the degradation rate is relatively high.
We believe this is because the SOTA model is more optimally tuned for offline evaluations and thus becomes relatively poorer for online incremental purposes. 
The proposed model is able to reset the states and effectively switch the intent outputs dynamically over sequences of utterances without undergoing much degradation in performance, thereby validating its use for incremental online SLU.
An important observation is that the model trained on utterance sequence length of 3 performs just as well, generalizing to higher length sequences.
Thus, we only consider online models trained on utterance sequence length of 3 from here onwards.

\begin{figure}[t]
\centering
\includegraphics[width=\columnwidth]{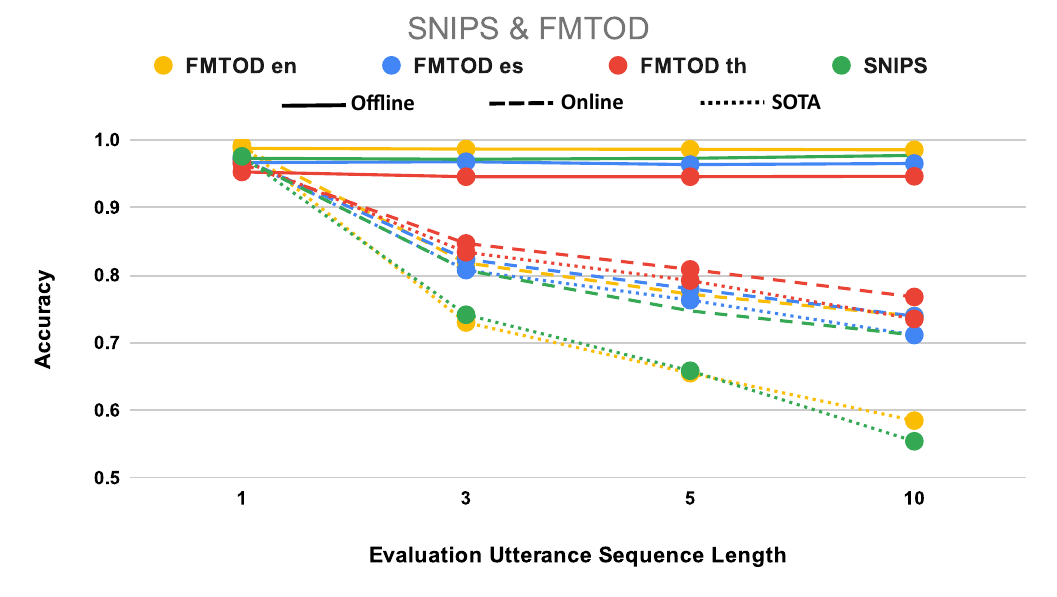}
\caption{\small{Accuracy of offline, SOTA vs. proposed online models at oracle EOS on Snips and FMTOD datasets\\
FMTOD en: English; FMTOD es: Spanish; FMTOD th: Thai}}\label{fig:snips_results}
\end{figure}

Fig~\ref{fig:pre_results} (see dotted lines) also plots the results obtained by evaluating models (trained using human annotated transcripts) on erroneous ASR transcripts.
As expected the errors in the transcripts cause the accuracy to fall considerably across all the models.
We believe the proposed models are effected two folds by the presence of errors: (i) directly on the intent prediction performance, and (ii) through EOS prediction, which in turn also effects the intent prediction process.
However, the results portray similar advantages as observed in section~\ref{sec:results_atis}.
The proposed systems are capable of maintaining high accuracy on ASR transcripts over higher utterance sequence lengths, while maintaining relatively close performance to baseline offline system for utterance sequence length of 1.
Moreover, the proposed system on ASR transcripts outperform both the state-of-the-art and baseline offline systems evaluated on clean manual transcripts for utterance sequence lengths greater than 1.

\begin{figure*}[t]
\centering
\includegraphics[width=\textwidth]{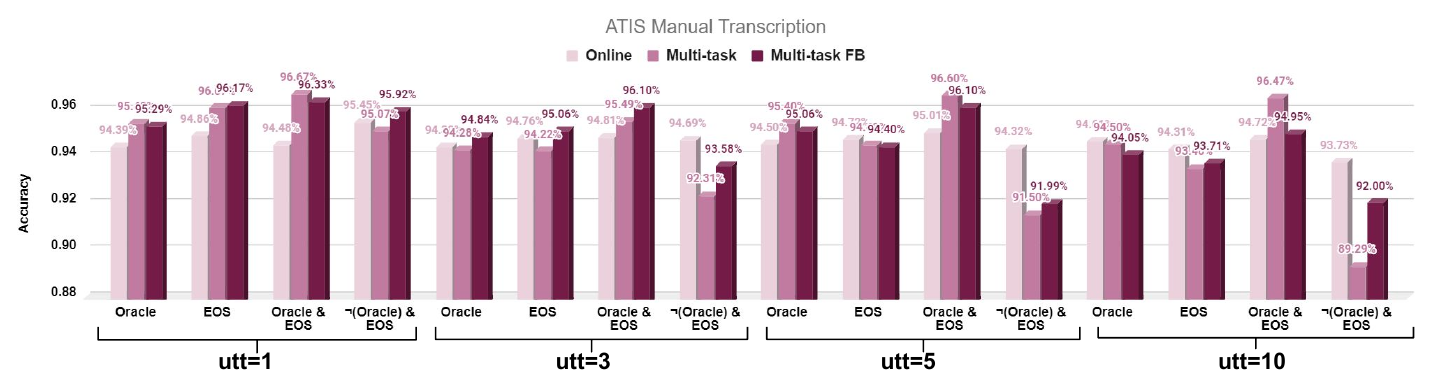}
\caption{Accuracy of three proposed online models evaluated on oracle and predicted sentence boundaries\\
Oracle: Intent accuracy evaluated at oracle EOS boundaries; EOS: Intent accuracy evaluated at predicted EOS boundaries;\\ Oracle \& EOS: Intent accuracy evaluated only at True positives of predicted EOS boundaries; $\lnot$(Oracle) \& EOS: Intent accuracy evaluated at False positives of predicted EOS boundaries.
}\label{fig:results}
\end{figure*}

\begin{figure*}[t]
\centering
\includegraphics[width=\textwidth]{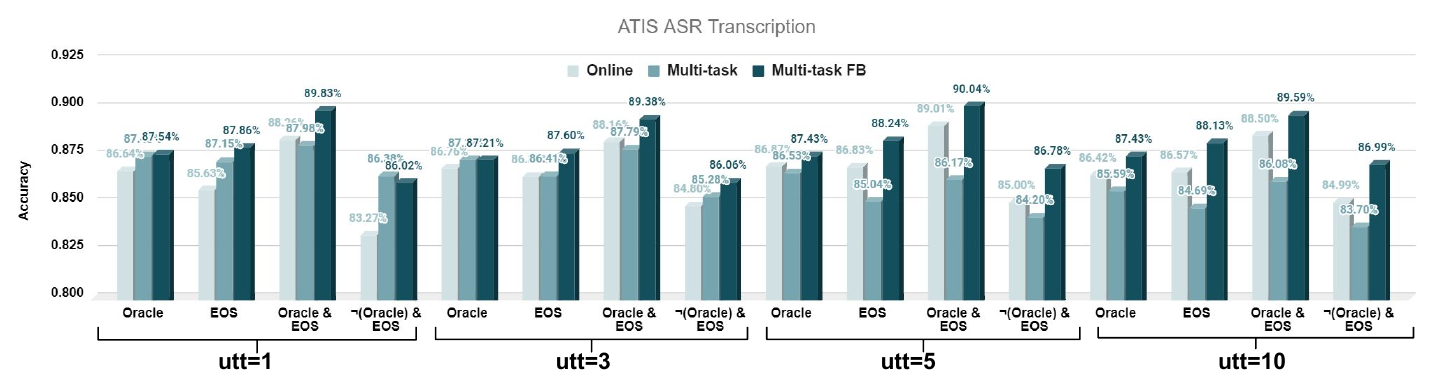}
\caption{Accuracy of three proposed online models evaluated on ASR transcripts at oracle and predicted sentence boundaries\\
Oracle: Intent accuracy evaluated at oracle EOS boundaries; EOS: Intent accuracy evaluated at predicted EOS boundaries;\\ Oracle \& EOS: Intent accuracy evaluated only at True positives of predicted EOS boundaries; $\lnot$(Oracle) \& EOS: Intent accuracy evaluated at False positives of predicted EOS boundaries.}\label{fig:asr_results}
\end{figure*}

\subsection{Transferability over Multiple Databases}
Fig~\ref{fig:snips_results} plots the results obtained on the Snips and FMTOD databases.
For Snips database we also compare with the SOTA system presented in \cite{e-etal-2019-novel}.
Comparing the SOTA and offline systems with the proposed model, the observations made are identical to those made with ATIS database in section~\ref{sec:results_atis} for both Snips and FMTOD databases.
This proves the efficacy of the proposed system over different databases, multiple domains, multiple languages and different vocabulary sizes.
We observe different degrees of degradation among the three languages in FMTOD databases.
In our experiments, we found the English language (Snips and FMTOD-en) undergoes high degradation whereas the Thai language (FMTOD-th) the least.
We attribute this to the structure and the linguistic characteristics of the languages.

\subsection{EOS Predictions}\label{sec:eos_pred}
Our experiments found that the EOS system performs consistently over varying utterance sequence lengths, achieving an accuracy of approximately 92\% with no significant performance differences between discussed architectures.

On using ASR outputs, the EOS prediction accuracy drops to approximately 89\%.
As earlier, no significant performance differences were noted across different architectures.

\subsection{EOS Evaluations}
Further, evaluations of the online intent module is performed by replacing the oracle EOS with predicted outputs of the EOS module.
With this, there are possibilities of false positives / negatives during EOS detection, hence, we also compute the accuracy of the intent classification when EOS predictions match with Oracle EOS (Oracle\&EOS).
The results of three proposed models are presented in Fig~\ref{fig:results} in terms of accuracies over varying utterance sequence lengths.
Note, for the online system without multi-task architecture, the results are evaluated by running the independent EOS detection system to obtain the estimated EOS boundaries.
Comparing with the offline version (see Fig~\ref{fig:pre_results}), the online version outperforms by a large margin evaluated on predicted EOS boundaries, except for utterance sequence length of 1 (offline evaluation).
Moreover, whenever the EOS predictions match with Oracle EOS (see Oracle\&EOS in Fig~\ref{fig:results}), the performance of the proposed systems are particularly high which is encouraging.
This underlines the practicality of proposed online SLU in conjunction with EOS predictor for incremental online processing.

Evaluations on the ASR transcripts are presented in Fig~\ref{fig:asr_results}.
The results portray similar observations as earlier, i.e., even when evaluating the intent predictions at predicted EOS boundaries, the performance of the proposed systems have drastic advantages over the baseline and SOTA systems (except for offline evaluation).


\begin{figure*}[t]
\centering
\begin{tikzpicture}
\node[anchor=south west, inner sep=0] at (0,0) {
\includegraphics[width=\textwidth]{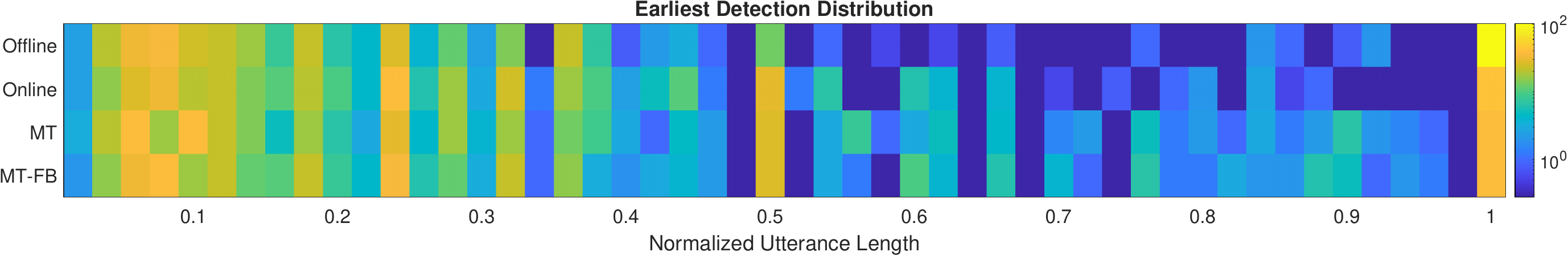}
};
\begin{axis}[at={(5,350)},domain=0:1,restrict y to domain=0:5,x=16.3cm,y=2.08cm,xmin=0,xmax=1,ymin=0,ymax=0.8,axis line style={draw=none},tick style={draw=none},ticks=none]
\addplot[smooth,color=red,mark=x,mark options={scale=1.2},line width=1.2pt]
  plot coordinates {
	  (0.279,0.1)
		(0.269,0.33)
		(0.281,0.56)
		(0.316,0.8)
	};
	\end{axis}
\end{tikzpicture}
\caption{\small{Early Detection Distribution: Offline: Baseline Offline RNN; Online: Proposed Online RNN + Independent EOS Detection; MT: Proposed Online Multi-task RNN; MT+FB: Proposed Online Multi-task RNN with Feedback;\\ 
Normalized Utterance Length of 0 refers to the beginning of the sentence and 1.0 refers to EOS; Color map indicate the counts of correct early predictions; Red markings are averages;}}\label{fig:early_dist}
\end{figure*}

\subsection{Multi-task Frameworks}
From Fig~\ref{fig:results} (Oracle, EOS and Oracle\&EOS), it is apparent that the multi-task frameworks provide better accuracies for intent detection.
However, the EOS detection performance is similar to the independent EOS system.
Comparing the vanilla multi-task model and the feedback version, the feedback version achieves better accuracies for utterance sequence lengths of 1 and 3.
We believe this is because the feed-back version is more sensitive to training data which was limited to utterance sequence lengths of 3.

However, with the evaluation on ASR outputs presented in Fig~\ref{fig:asr_results} the advantages of the multi-task model with EOS feedback is much more evident.
We find that the multi-task model with EOS feedback consistently outperforms the other architectures for all the utterance sequence lengths.
We believe this is because the system relies on EOS information more in the presence of the noise (errors) in ASR transcripts.
And since the EOS predictor is more robust to errors in the transcript (see section~\ref{sec:eos_pred}: an absolute drop of approximately 3\% with EOS predictions versus $>$ 6\% absolute drop in performance with intent detection), it is able to provide the crucial information required to switch intent states at utterance boundaries.

\subsection{Early Prediction of Intents}
One of the advantages of the online SLU is the potential to predict the intent prematurely.
To evaluate the prospects of this, we calculate the accuracy of intent predictions during false positives of the EOS detection (see $\lnot$(Oracle)\&EOS in Fig~\ref{fig:results} and Fig~\ref{fig:asr_results}).
We observe that the accuracy of intent detection at false positives to be high and close to the accuracy at Oracle EOS.
This implies that the intent detection is accurate even before the EOS, thereby hinting at possibilities of early prediction.

Additionally, we analyze the earliest time (in terms of number of words) the network starts predicting the correct intent.
We obtain the early detection distribution normalizing over utterance length and then over class distribution (see Fig~\ref{fig:early_dist}).
The x-axis corresponds to normalized utterance length (0.0 is start of utterance; 1.0 is EOS) and the heat-map correspond to the number of utterances.
We observe an acute peak at 1.0 for offline system, suggesting the majority of intents are detected correctly only at EOS.
However, for the proposed system, the peak at 1.0 is less pronounced and the distribution is concentrated relatively more towards smaller values, thus suggestive of earlier detections.
We find that the early detections with our proposed models are statistically significant ($p$-value$<$0.005) compared to offline system.
Moreover, the proposed multi-task model is capable of earlier detections (significant, $p$-value$<$0.05) than the online model.


\section{Conclusion}
\label{sec:conclusion}
In this paper, we motivate the need for incremental and online processing for SLU.
The low latency profile, prospects of early intent detection, independence from ASR end-pointing makes the approach attractive.
We define the incremental online SLU as real-time spoken intent inference on a continuous streaming sequence of utterances with no sentence demarcation.
We demonstrate that the typical offline approaches to SLU are unsuitable for incremental online processing.
Multiple approaches to online SLU are proposed based on RNN intent classification.
For determining the sentence boundaries, EOS detection is performed.
Multi-task learning is proposed to better model the EOS and the intent detection jointly.
Proposed models enable switching of intent states implicitly, accompanying estimation of sentence boundaries.
Final results of the proposed techniques are indicative of performance approaching the accuracies of offline SLU.
Sustained advantages on erroneous offline ASR transcripts show promise towards real-world application.

In the future, we would like to make evaluations on top of a real-time online ASR both in terms of latency profiling and accuracy profiling.
The impact of ASR errors for offline versus the proposed system is worthy of investigation.
Most importantly, we would like to compare the performance between the ASR end-pointing based SLU versus the proposed incremental processing.
Finally, we intend to employ more complex neural network architectures, include character encodings and extend the framework to other SLU tasks like slot-filling, named-entity detection and assess the impact of incremental online processing.
Developing heuristics for early prediction of spoken language intent is also an area of interest.
Different weighting schemes on the loss function can be explored. 
Particularly, giving less importance to the intent at the beginning of the utterance and increasing weights along the sentence length could make the training more efficient. 
Conversely, providing more weight towards the start of the sentence could facilitate early detection.

\bibliographystyle{IEEEbib}
\bibliography{refs,refs2}

\end{document}